%% file: root.tex
\documentclass[runningheads]{llncs}

\usepackage{eccv}

\usepackage{eccvabbrv}
\usepackage{graphicx}
\usepackage{booktabs}
\usepackage[accsupp]{axessibility}

\usepackage{amsmath}
\usepackage{array}
\usepackage{rotating}
\usepackage{placeins}

\usepackage{hyperref}

\usepackage{orcidlink}

\title{BikeScenes: LiDAR Semantic Segmentation for Bicycles}
\titlerunning{BikeScenes}

\author{Denniz Goren\inst{1} \and
Holger Caesar\inst{1}\orcidlink{0000-0001-5099-6297}}
\authorrunning{D. Goren and H. Caesar}
\institute{Department of Cognitive Robotics, Delft University of Technology, Delft, The Netherlands\\
\email{dennizgoren@gmail.com, h.caesar@tudelft.nl}}

\begin{document}
\maketitle

\input{Chapters/Abstract}
\input{Chapters/Introduction}
\input{Chapters/Related_Work}
\input{Chapters/BikeScenes}
\input{Chapters/Experiments}
\input{Chapters/Conclusion}

\bibliographystyle{splncs04}
\bibliography{bibliography}

\end{document}

%% file: Chapters/Abstract.tex
\begin{abstract}
\label{abstract}
The vulnerability of cyclists, exacerbated by the rising popularity of faster e-bikes, motivates adapting automotive perception technologies for bicycle safety. We use our multi-sensor SenseBike research platform to study 3D LiDAR semantic segmentation for bicycles. We introduce the novel BikeScenes-\allowbreak lidarseg Dataset, comprising 3021 consecutive LiDAR scans around the university campus of the TU Delft, semantically annotated for 29 dynamic and static classes. As an initial baseline study, we evaluate how a Semantic\-KITTI pre-trained FRNet model transfers to this bicycle-mounted solid-state LiDAR setting. Fine-tuning on BikeScenes increases mean Intersection-over-Union (mIoU) from 13.8\% to 63.6\% on our held-out subsequences. These results show the practical value of in-domain data for this platform, while also highlighting the need for larger bicycle datasets. We contribute BikeScenes as a resource for advancing research in cyclist-centric perception.
\keywords{LiDAR semantic segmentation \and Bicycle perception \and Transfer learning \and Dataset}
\end{abstract}

%% file: Chapters/Introduction.tex
\section{Introduction}
\label{sec:introduction}

The inherent vulnerability of cyclists presents a persistent safety challenge, particularly in regions like the Netherlands, where cycling is vital to urban mobility. This situation is further complicated by the integration of faster e-bikes, introducing new potential risks that demand attention. While advanced driver-assistance systems (ADAS), powered by sensors and the integration of machine learning, are significantly enhancing safety in automobiles, similar technologies for bicycles are far less developed. With sensors such as LiDAR becoming more compact and affordable, the question arises: can we leverage these advancements to improve bicycle safety by creating assistive systems with precise awareness of their complex surroundings?

We focus on LiDAR because it provides direct 3D geometric information about the surrounding scene and is less dependent on image appearance and lighting than camera-only perception. This makes it useful for mapping, obstacle awareness, and semantic scene understanding from a bicycle platform. The camera on the SenseBike is therefore used as supporting information during annotation, while the segmentation task in this work is defined on LiDAR point clouds.

\begin{figure}[htbp]
    \centering
    \includegraphics[width=0.94\linewidth]{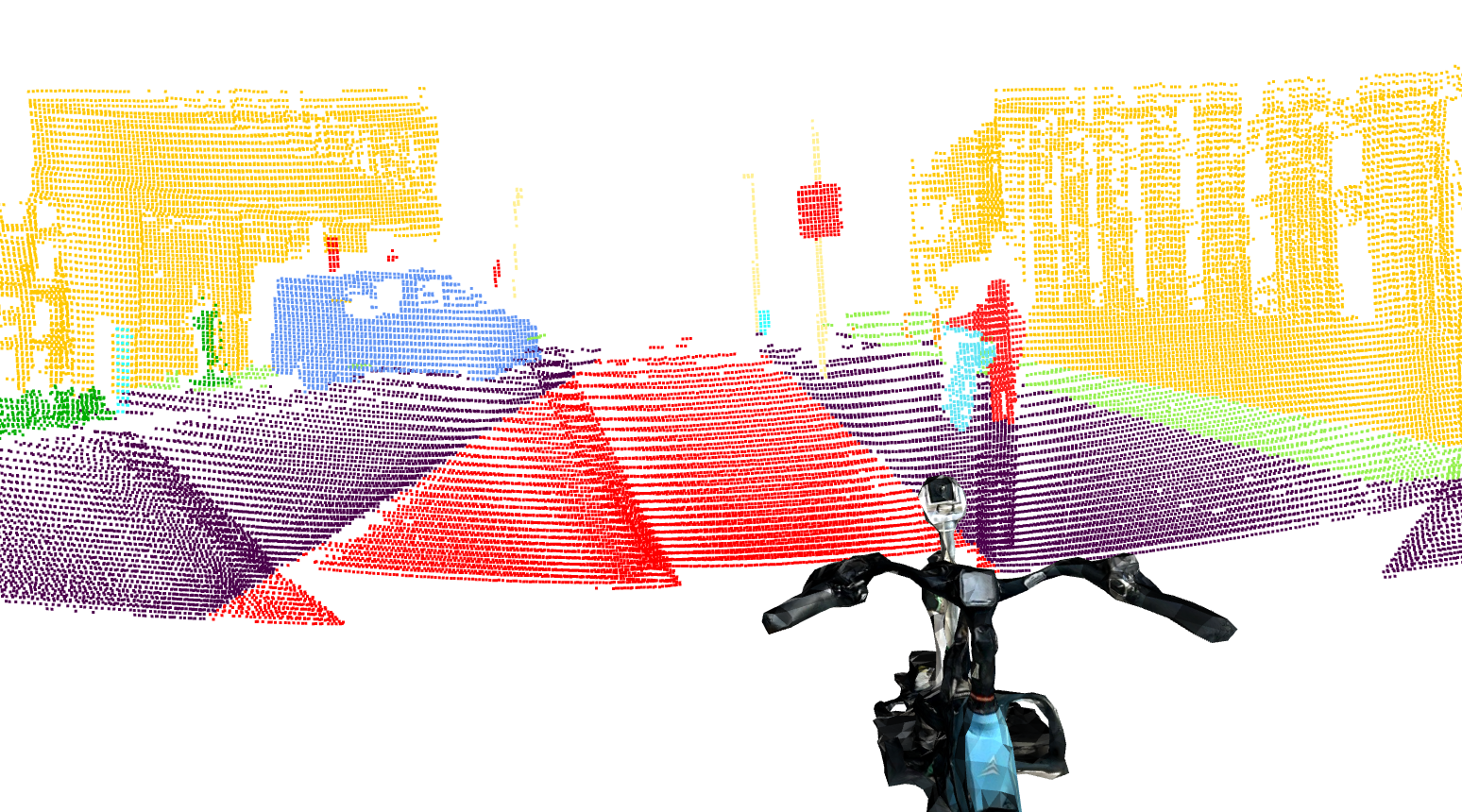}
    \caption{Scene from the BikeScenes-lidarseg Dataset.}
    \label{fig:openbike_scene}
\end{figure}

This work explores this question by developing and evaluating a 3D LiDAR semantic segmentation system specifically designed for a bicycle platform. Accurate environmental understanding--identifying infrastructure, road users, and other obstacles--is an essential prerequisite for any potential bicycle ADAS or enhanced rider awareness system. To facilitate this research, we utilize the SenseBike, a research platform based on the Boreal~Holoscene~X~\cite{borealbikes}, equipped with a sensor suite including LiDARs, an IMU, GPS, cameras, and an NVIDIA Jetson for onboard computation. 

Implementing such a system on a bicycle introduces unique challenges, such as calibrating sensors on a highly dynamic platform, correcting motion-induced scan skew, and adapting perception models trained on automotive data to a bicycle-mounted sensor setup. To address this, we introduce a dataset collected from a bicycle's perspective: \textbf{BikeScenes-lidarseg}. It captures classes and scene geometry underrepresented in automotive benchmarks. We further provide an initial baseline study with FRNet to quantify the shift between Semantic\-KITTI pre-training and our bicycle-mounted LiDAR data.

%% file: Chapters/Related_Work.tex
\section{Related Work}
\label{sec:related_work}

\subsection{Outdoor LiDAR Segmentation Datasets}
\label{subsec:datasets}

Large-scale automotive datasets such as Semantic\-KITTI~\cite{SemanticKITTI}, nuScenes~\cite{nuscenes}, and Waymo~\cite{Waymo} are today's primary publicly available benchmarks for 3D semantic segmentation. Although they have driven significant progress in the field, these datasets are captured from sensors mounted on the roofs of cars. As a result, their direct transfer to a bicycle-mounted setup is affected by differences in perspective and positioning on the road. This shift is further exacerbated by differences in sensor characteristics, including field-of-view, intensity distributions, and scan-line patterns. The Salzburg Bicycle LiDAR Dataset~\cite{salzburg} was the first bicycle-centric LiDAR segmentation dataset, focusing primarily on static environmental classes such as roads, buildings, and vegetation. BikeScenes complements this work with annotations for dynamic road users and a labeling scheme aligned with Semantic\-KITTI for baseline transfer experiments.

\subsection{LiDAR Segmentation Architectures}
\label{subsec:segmentation_architectures}
Deep learning methods dominate the task of LiDAR semantic segmentation. Common approaches process the irregular point cloud structure by operating directly on points~\cite{qi2017pointnet, thomas2019kpconv}, using voxel representations~\cite{zhu2020cylindrical, pvkd}, projecting onto 2D range views~\cite{xu2025frnet, cheng2022cenet}, or combining point and voxel information~\cite{SPV}. Recently, Transformer-based models have set new benchmarks by leveraging self-attention to capture global context~\cite{PT, PTV2, PTV3, SphereFormer}. A key challenge, however, for online applications such as the SenseBike is the trade-off between the high accuracy of state-of-the-art models~\cite{xu2025frnet, PTV3, SphereFormer} and the computational efficiency required for deployment on resource-constrained hardware.

%% file: Chapters/BikeScenes.tex
\section{BikeScenes Lidarseg Dataset}
\label{sec:openbike}
Public datasets such as Semantic\-KITTI~\cite{SemanticKITTI} and nuScenes~\cite{nuscenes} have become cornerstone benchmarks for 3D automotive perception. However, their direct applicability to bicycle-centric perception is limited: they are recorded using car-roof sensors in Germany and the USA/Singapore and differ from SenseBike's data in sensor type and placement, viewing angle, field-of-view, and scene content. To evaluate the benefit of in-domain data, we created \textbf{BikeScenes-lidarseg} from the SenseBike's perspective. It follows the Semantic\-KITTI labeling scheme to allow direct comparison, while adding a separate \emph{bike-path} class for future cyclist-specific experiments.

\begin{table}[htbp]
\centering
\footnotesize
\renewcommand{\arraystretch}{1}
\setlength{\tabcolsep}{6pt}
\caption{Comparison of outdoor LiDAR semantic-segmentation datasets.}
\label{tab:dataset_comparison}
\resizebox{\textwidth}{!}{%
\begin{tabular}{lcccccccccc}
\toprule
\textbf{Dataset} 
  & \textbf{Platform}      
  & \textbf{Countries}       
  & \textbf{Sequences}      
  & \textbf{Ann.\ Frames}   
  & \textbf{Classes}        
  & \textbf{LiDAR}                       \\
\midrule
SemanticKITTI \cite{SemanticKITTI}
  & Car 
  & Germany 
  & 22 
  & 43,552 
  & 28 
  & Velodyne HDL-64E   \\
nuScenes \cite{nuscenes}
  & Car 
  & USA, Singapore
  & 1,000 
  & 40,000
  & 32
  & Velodyne HDL-32E   \\
Waymo \cite{Waymo}
  & Car 
  & USA
  & 1,150
  & 46,000
  & 23
  & Undisclosed 64-beam rotating \\
SemanticPOSS \cite{pan2020semanticposs}
  & Car 
  & China
  & 1
  & 2,988
  & 14
  & Hesai Pandora  \\
PandaSet \cite{pandaset}
  & Car 
  & USA
  & 30
  & 6,080
  & 37
  & Pandar64 + PandarGT \\
SBLD \cite{salzburg}
  & Bicycle 
  & Austria
  & 17
  & 9,486
  & 10
  & 5 x Livox Horizon  \\
BikeScenes-lidarseg  
  & Bicycle 
  & the Netherlands
  & 1
  & 3,021
  & 29
  & RoboSense M1 Plus  \\
\bottomrule
\end{tabular}
}
\end{table}

\subsection{Recording}
\begin{figure}[!ht]
    \centering
    \includegraphics[width=0.95\linewidth]{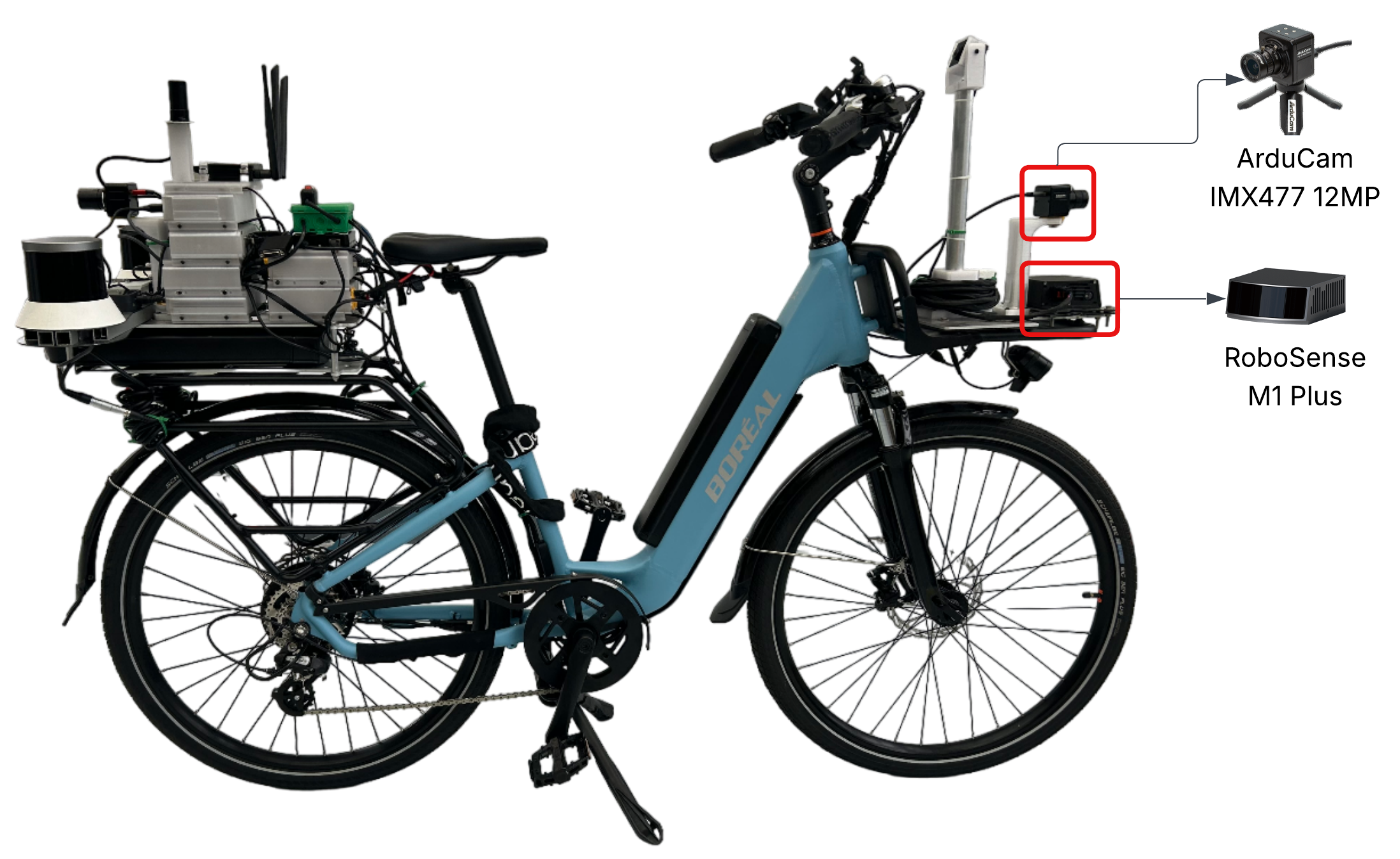}
    \caption{The SenseBike and sensors used to capture BikeScenes.}
    \label{fig:sensebike}
\end{figure}

As shown in Figure~\ref{fig:sensebike}, a RoboSense M1 Plus LiDAR mounted on the front carrier of the SenseBike was used to record a 1.3~km closed loop at 10~Hz, yielding 3021 consecutively time-stamped scans around a university campus. The scans were online compensated for ego motion, using GPS and IMU-based odometry. Additionally, a front-facing camera captured images at 10Hz to aid in the labeling procedure. 

The LiDAR frames were registered offline with GLIM~\cite{Koide_2024} to construct the aggregated point-cloud map used for multi-scan labeling. Table~\ref{tab:dataset_comparison} compares BikeScenes with other related outdoor LiDAR segmentation datasets.

\subsection{Labeling Procedure}  
\begin{figure}[!ht]
  \centering
  \includegraphics[width=0.75\linewidth]{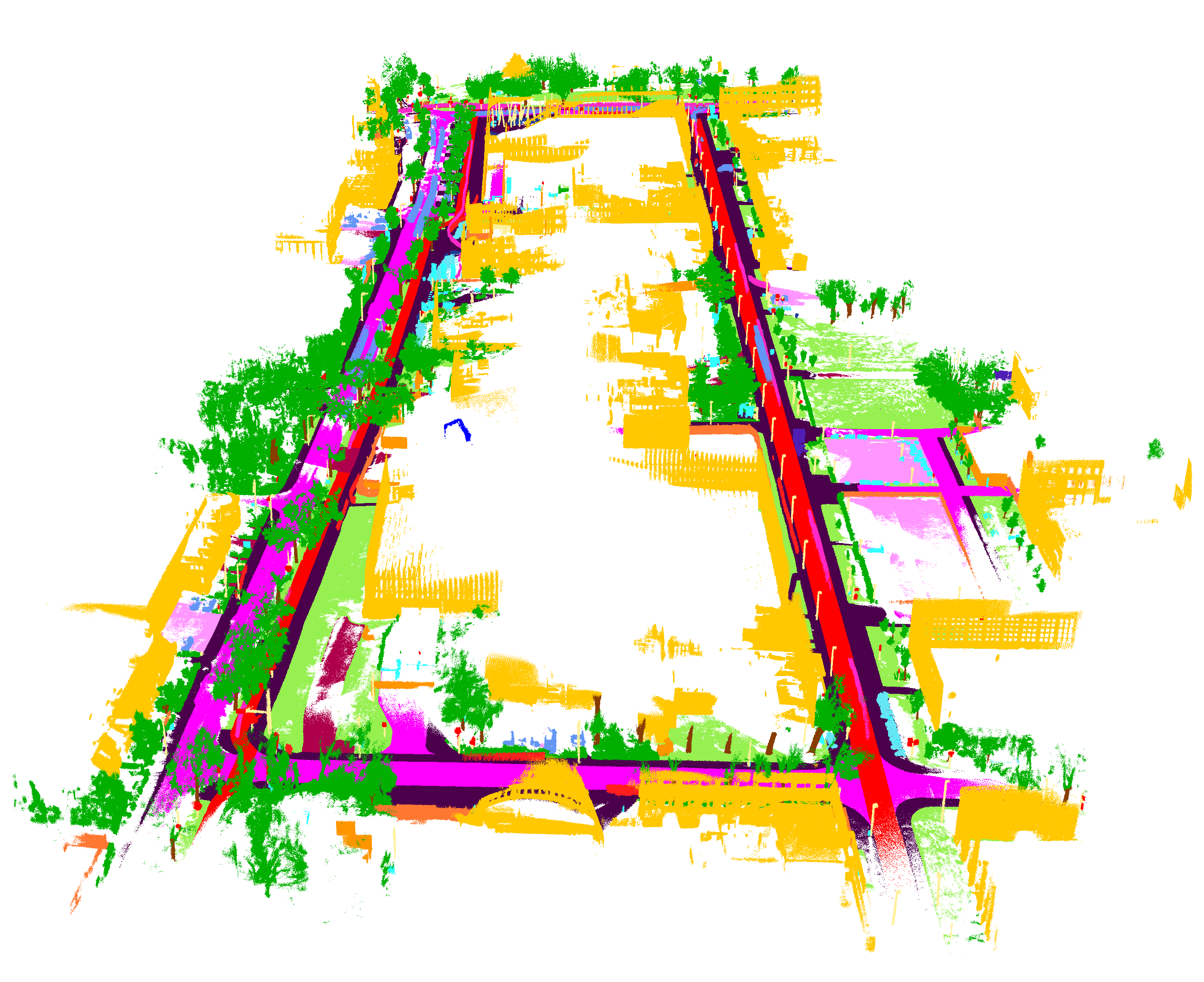}
  \caption{Map of aggregated labeled scans.
    \textcolor[RGB]{255,200,0}{\rule{0.8em}{0.8em}} building;  
    \textcolor[RGB]{255,0,255}{\rule{0.8em}{0.8em}} road;
    \textcolor[RGB]{75,0,75}{\rule{0.8em}{0.8em}} sidewalk;  
    \textcolor[RGB]{0,175,0}{\rule{0.8em}{0.8em}} vegetation;  
    \textcolor[RGB]{255,0,0}{\rule{0.8em}{0.8em}} bike-path.
  }
  \label{fig:map}
\end{figure}

\begin{figure}[!t]
    \centering
    \includegraphics[width=0.90\textwidth]{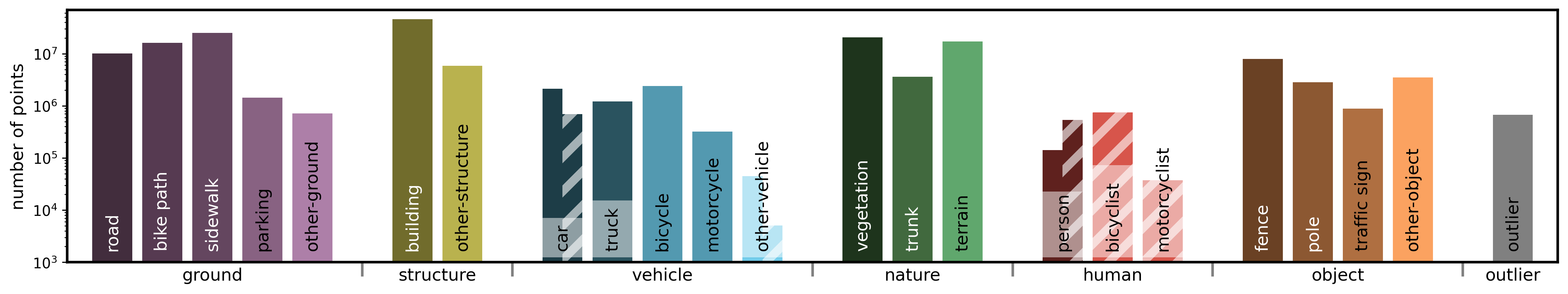}
    \caption{Class distribution of the BikeScenes dataset. The number of points for dynamic classes is divided between non-moving (solid bars) and moving objects (hatched bars).}
    \label{fig:class_distribution}
\end{figure}

All annotations were produced by a single annotator using the SemanticKITTI Labeling Tool and a consistent labeling protocol. Consequently, inter-annotator agreement was not assessed. We preserved the original 28 classes (22 unique + 6 moving/static versions) and added a \emph{bike-path} class.
Static objects were labeled in the aggregated map, while dynamic objects (pedestrians, cyclists, cars) were labeled \emph{per scan} to ensure high data quality for these sparse and fine-scale classes, of which the structure is represented by only a small number of points. The labeled map consisting of all the scans combined is shown in Figure \ref{fig:map}. The class distribution of BikeScenes is provided in Figure \ref{fig:class_distribution}.

\FloatBarrier

\subsection{Subsequences - Train/Val/Test Split}
\begin{figure}[!t]
    \centering
    \includegraphics[width=0.92\linewidth,trim=0 12.82bp 0 12.82bp,clip]{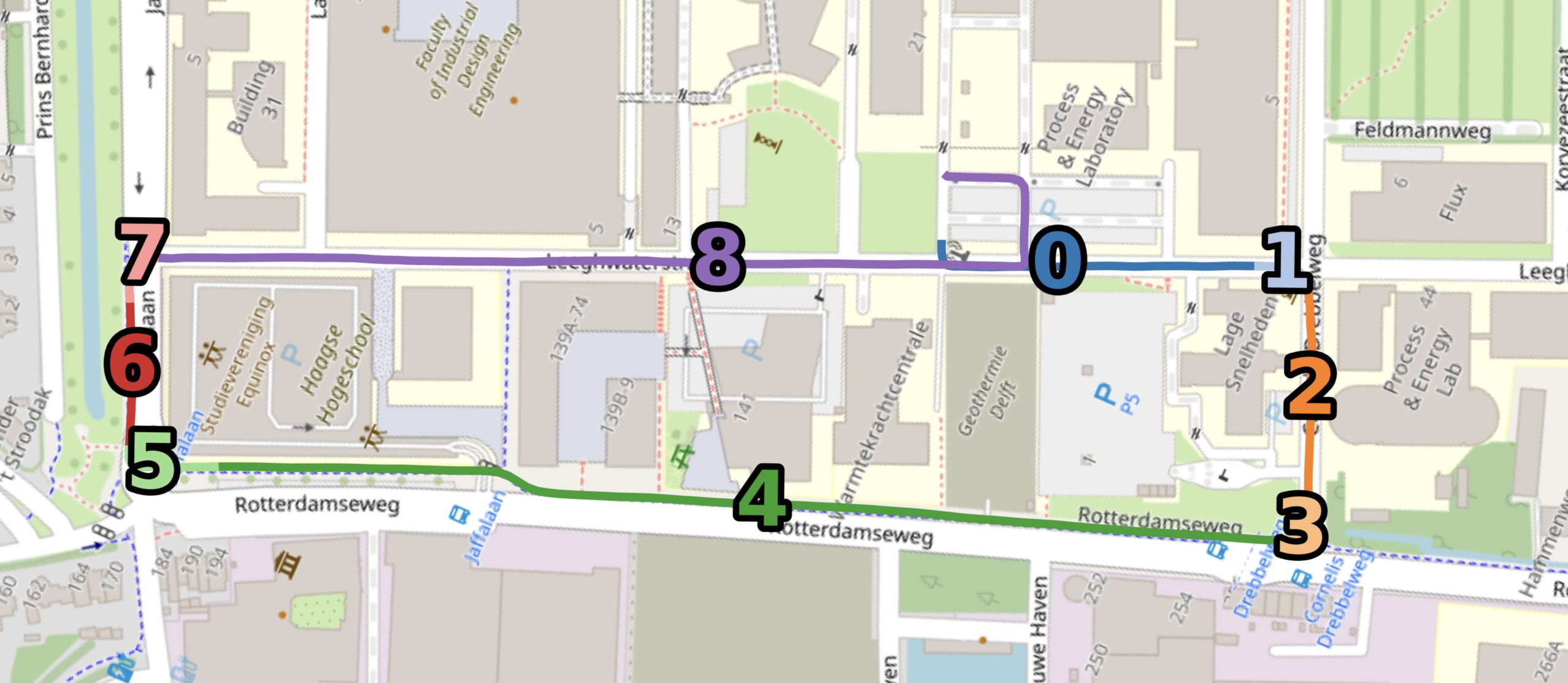}
    \caption{GPS trajectory and subsequence categorization. Basemap \copyright{} OpenStreetMap contributors~\cite{openstreetmap}, available under the Open Database License.}
    \label{fig:subsequences}
\end{figure}

We divided the continuous recording into nine spatial subsequences, as shown in Figure~\ref{fig:subsequences}. The training set contains the three long sections 0, 4, and 8 (2,358 scans), the test set contains sections 2 and 6 (358 scans), and the four intervening corner sections 1, 3, 5, and 7 form the validation set (305 scans).

The split boundaries were selected by visually inspecting the recording and placing them where the following straight section was occluded by the geometry of the corner. Consequently, each test section is separated from the training sections by corner subsequences assigned to validation. These intervening sections contain between 57 and 112 scans and reduce direct spatial and temporal overlap between the training and test sets.

\paragraph{Data and code availability.}
The BikeScenes-lidarseg dataset is publicly archived on Zenodo under the CC BY 4.0 license~\cite{goren2025bikescenesdataset}. Dataset documentation, split definitions, and visualization tools are available in the accompanying GitHub repository~\cite{bikescenesrepository}.

%% file: Chapters/Experiments.tex
\section{Experiments and Results}
\label{sec:experiments}
To measure the effect of training on BikeScenes and support deployment on the SenseBike, we first select an architecture that balances accuracy and inference speed on the onboard Jetson Orin NX. We then compare Semantic\-KITTI pre-training with training and fine-tuning on BikeScenes, and evaluate performance on the held-out test split.

\subsection{Evaluation Metrics}
For our experiments, we use the standard intersection-over-union (IoU) metric for semantic segmentation to evaluate the performance of the different models and configurations quantitatively. 

For a given class $i$, IoU is defined as: 
\begin{equation}
IoU_i = \frac{TP_i}{TP_i + FP_i + FN_i}.
\end{equation}

where $TP_i$ and $FP_i$ are the true and false positives, respectively, and $FN_i$ the false negatives for class $i$. The mean over all the evaluated classes is often calculated and presented as $mIoU$.

\subsection{Architecture Selection for On-bike Inference}
To evaluate own-domain training on BikeScenes and run segmentation on the SenseBike, the selected architecture must combine segmentation quality with inference efficiency. Direct comparison of author-reported FPS is difficult because methods are evaluated on different GPUs. We therefore use normalized FPS as a coarse efficiency metric:

\begin{equation}
\text{normalized FPS} =
\frac{\text{author-reported FPS}}{\text{GPU peak FP32 TFLOPS}}.
\end{equation}

\begin{figure}[t]
    \centering
    \includegraphics[width=0.68\linewidth]{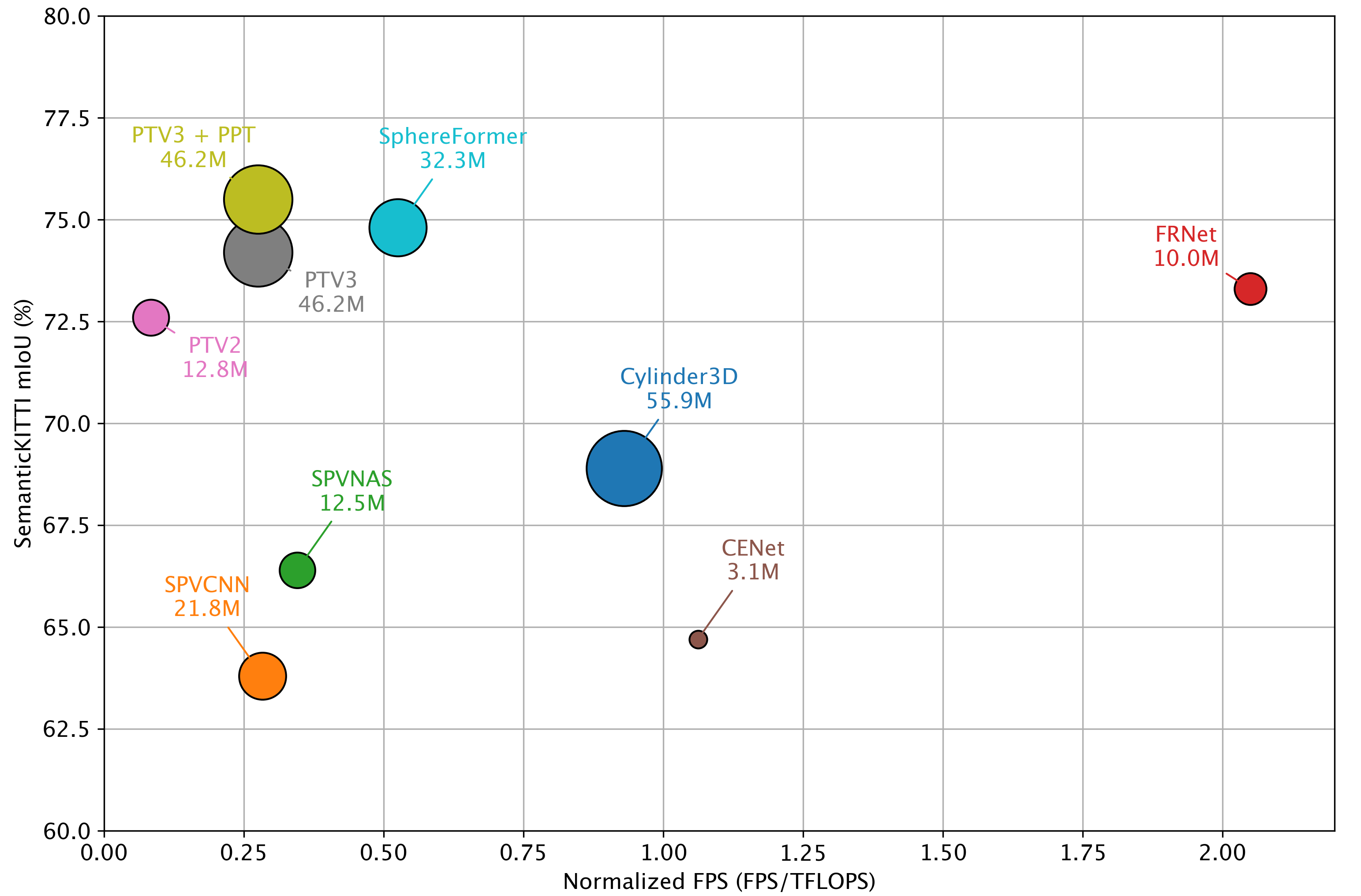}
    \caption{Semantic\-KITTI test-set mIoU versus normalized FPS. Marker size represents the number of model parameters.}
    \label{fig:normalized_fps}
\end{figure}

Figure~\ref{fig:normalized_fps} compares the official Semantic\-KITTI test-set mIoU with normalized FPS, using the throughput and hardware reported by the respective authors~\cite{zhu2020cylindrical,SPV,xu2025frnet,cheng2022cenet,PTV2,PTV3,PPT,SphereFormer}. Although this normalization does not replace benchmarking all models on the same device, it provides an initial comparison across the reported hardware.

FRNet~\cite{xu2025frnet} achieves 73.3\% mIoU on Semantic\-KITTI with a normalized FPS of 2.05, compared with the highest mIoU of 75.5\% at 0.27 normalized FPS for PTV3 with PPT~\cite{PTV3,PPT}. Given this balance and its lightweight design, we use FRNet for all subsequent experiments on BikeScenes.

\subsection{Class Remapping}
For the following experiments, the 29 classes of the BikeScenes dataset are remapped to the 19 Semantic\-KITTI evaluation classes. We follow the Semantic\-KITTI evaluation convention by collapsing moving/static variants into their corresponding evaluation classes and ignoring labels outside the 19 evaluated classes. The released BikeScenes annotations retain \emph{bike-path} as a separate class for future cyclist-specific experiments. However, Semantic\-KITTI has no corresponding output class; we therefore merge \emph{bike-path} into \emph{road} only for the Semantic\-KITTI-compatible baseline experiments reported here. This alignment enables evaluation using Semantic\-KITTI-trained models without introducing a new output class. All training and testing configurations in this paper use this 19-class mapping.

Although all configurations use the same 19-class output mapping, not every evaluation class is represented in each split. The validation split contains 17 evaluation classes, with \emph{truck} and \emph{other-vehicle} absent. The test split contains 16 evaluation classes, with \emph{truck}, \emph{other-vehicle}, and \emph{motorcyclist} absent. We calculate mIoU over the classes present in each split. Consequently, validation and test mIoU are based on different present-class subsets and are not intended as a direct cross-split comparison.

\subsection{Baseline Transfer and Training Schedules}
We first adopt the FRNet schedule reported by the authors (One-Cycle~\cite{smith2018superconvergencefasttrainingneural}, 50k iters, peak learning rate (LR) $0.01$, batch size 2) and evaluate (i) pre-trained on Semantic\-KITTI only, (ii) training from scratch on BikeScenes, and (iii) fine-tuning on BikeScenes. We then repeat with a lower peak LR ($3{\times}10^{-4}$) and vary training length across 10k, 30k, and 50k iterations. All runs use the 19-class BikeScenes-to-Semantic\-KITTI mapping.

Table~\ref{tab:results} shows three consistent trends. First, the model trained on Semantic\-KITTI transfers poorly to BikeScenes, indicating a substantial shift between the automotive benchmark data and our bicycle-mounted solid-state LiDAR data. Second, on the validation split, fine-tuning on BikeScenes consistently outperforms training from scratch across matched schedules, showing that pre-trained weights still provide useful features. Third, lowering the peak LR to $3{\times}10^{-4}$ improves fine-tuning (but not from-scratch training), with performance peaking at 30k iterations and diminishing beyond. We therefore select Cfg.~2d (fine-tune, $3{\times}10^{-4}$, 30k) for testing.

\subsection{Performance on the Test Set} 
On the held-out test split, the validation-selected fine-tuned checkpoint (Cfg.~2d) increases mIoU from 13.8\% to 63.6\% over the Semantic\-KITTI-only baseline (Table~\ref{tab:results}). Scratch-trained configurations are compared only on validation. This gap confirms the benefit of in-domain fine-tuning and shows that automotive pre-training alone is insufficient for BikeScenes.

Per-class results are strong for static scene classes such as \emph{road} (87.2\%), \emph{sidewalk} (83.4\%), \emph{building} (91.6\%), and \emph{vegetation} (82.5\%), and key dynamic classes such as \emph{person} (83.2\%) and \emph{bicyclist} (87.5\%). For this initial baseline, the weakest classes remain \emph{bicycle} (26.9\%), \emph{motorcycle} (32.4\%), and \emph{parking} (3.4\%). This cannot be attributed to point counts alone, as they do not capture instance diversity, occlusion, object completeness, or fine-grained geometric ambiguity in LiDAR-only views (e.g., parking vs. road or sidewalk).

\begin{table}[t]
\centering
\footnotesize
\renewcommand{\arraystretch}{1.1}
\setlength{\tabcolsep}{1pt}
\caption{Class-wise IoU (\%) on BikeScenes. All rows use the same 19-class output mapping. Rows marked \emph{Val} use the validation split and report mIoU over its 17 present classes; \emph{truck} and \emph{other-vehicle} are absent. Rows marked \emph{Test} use the test split and report mIoU over its 16 present classes; \emph{truck}, \emph{other-vehicle}, and \emph{motorcyclist} are absent. Best per-column performance among validation configurations is shown in \textbf{bold}. Test rows are shown without bolding. Dashes denote classes absent from the corresponding split.}

\label{tab:results}
\resizebox{\textwidth}{!}{%
\begin{tabular}{c@{\hspace{4pt}}l@{\hspace{4pt}}c@{\hspace{4pt}}c@{\hspace{4pt}}c | c | *{19}{c}}
\toprule
\textbf{Cfg.} & \textbf{Strategy} & \textbf{Peak LR} & \textbf{Iter.} & \textbf{Split} &
\rotatebox{90}{mIoU (\%)} &
\rotatebox{90}{Car} & \rotatebox{90}{Bicycle} & \rotatebox{90}{Motorcycle}
& \rotatebox{90}{Truck} & \rotatebox{90}{Other-Veh.} & \rotatebox{90}{Person}
& \rotatebox{90}{Bicyclist} & \rotatebox{90}{Motorcyclist} & \rotatebox{90}{Road}
& \rotatebox{90}{Parking} & \rotatebox{90}{Sidewalk} & \rotatebox{90}{Other-Gnd.}
& \rotatebox{90}{Building} & \rotatebox{90}{Fence} & \rotatebox{90}{Vegetation}
& \rotatebox{90}{Trunk} & \rotatebox{90}{Terrain} & \rotatebox{90}{Pole}
& \rotatebox{90}{Traffic-Sign} \\
\midrule
1a & Pre-trained & -- & --    & Val & 12.2 & 3.6 & 0.0 & 0.0 & -- & -- & 0.1 & 0.0 & 0.0 & 26.5 & 0.5 & 1.6 & 0.0 & 43.2 & 1.9 & 27.0 & 2.3 & 21.7 & 35.1 & 43.5 \\
1b & Scratch     & 0.01 & 50k & Val & 53.3 & 32.6 & \textbf{85.1} & 30.2 & -- & -- & 68.0 & 59.1 & 0.0 & 86.6 & \textbf{5.4} & 74.6 & 1.9 & 86.7 & 38.9 & 71.6 & 55.3 & 67.8 & 66.9 & 76.5 \\
1c & Fine-Tune   & 0.01 & 50k & Val & 55.6 & 30.8 & 84.2 & \textbf{35.2} & -- & -- & 74.7 & 63.2 & 0.0 & 86.8 & 3.7 & 74.1 & 2.6 & 89.2 & 42.9 & 76.6 & 60.1 & \textbf{71.8} & 70.9 & 78.4 \\
2a & Scratch     & $3{\times}10^{-4}$ & 10k & Val & 46.2 & 16.4 & 73.2 & 22.5 & -- & -- & 57.9 & 54.6 & 0.0 & 75.6 & 1.1 & 66.2 & 0.2 & 81.6 & 27.6 & 63.4 & 47.9 & 63.9 & 60.4 & 73.2 \\
2b & Fine-Tune   & $3{\times}10^{-4}$ & 10k & Val & 55.9 & 47.8 & 75.4 & 21.9 & -- & -- & \textbf{77.3} & 63.6 & 0.0 & 84.7 & 2.1 & 75.0 & 0.6 & 90.9 & 47.5 & 74.8 & 70.6 & 66.8 & 70.0 & \textbf{81.1} \\
2c & Scratch     & $3{\times}10^{-4}$ & 30k & Val & 51.5 & 33.2 & 81.2 & 24.7 & -- & -- & 62.2 & 61.0 & 0.0 & 84.4 & 2.3 & 72.3 & 1.3 & 85.4 & 35.3 & 68.3 & 55.5 & 66.1 & 67.2 & 75.2 \\
2d & Fine-Tune   & $3{\times}10^{-4}$ & 30k & Val & \textbf{58.6} & \textbf{65.1} & 80.2 & 21.6 & -- & -- & \textbf{77.3} & 65.2 & 0.0 & \textbf{88.7} & 5.0 & \textbf{75.3} & 4.6 & \textbf{92.1} & \textbf{49.1} & 77.5 & \textbf{70.9} & 71.1 & \textbf{72.0} & 80.5 \\
2e & Scratch     & $3{\times}10^{-4}$ & 50k & Val & 52.7 & 32.9 & 83.6 & 29.7 & -- & -- & 63.4 & 61.4 & \textbf{0.1} & 85.7 & 2.6 & 71.7 & 3.0 & 85.5 & 41.3 & 70.0 & 56.1 & 66.5 & 67.0 & 75.6 \\
2f & Fine-Tune   & $3{\times}10^{-4}$ & 50k & Val & 57.8 & 60.7 & 83.5 & 22.2 & -- & -- & 74.9 & \textbf{65.5} & 0.0 & 87.3 & 4.4 & 73.1 & \textbf{5.6} & \textbf{92.1} & 48.0 & \textbf{77.9} & 66.5 & 69.9 & 71.8 & 79.0 \\
\midrule
1a & Pre-trained & -- & --    & Test & 13.8 & 12.3 & 0.0 & 0.0 & -- & -- & 0.0 & 0.0 & -- & 44.1 & 0.6 & 1.5 & 0.5 & 45.1 & 1.4 & 24.7 & 1.4 & 26.1 & 29.7 & 33.0 \\
2d & Fine-Tune   & $3{\times}10^{-4}$ & 30k & Test & 63.6 & 78.9 & 26.9 & 32.4 & -- & -- & 83.2 & 87.5 & -- & 87.2 & 3.4 & 83.4 & 2.6 & 91.6 & 79.4 & 82.5 & 61.6 & 67.2 & 72.0 & 78.5 \\
\bottomrule
\end{tabular}
}
\end{table}

\subsection{Qualitative Results}
\begin{figure}[htbp]
    \centering
    \begin{minipage}{0.325\textwidth}
        \centering
        \includegraphics[width=\linewidth]{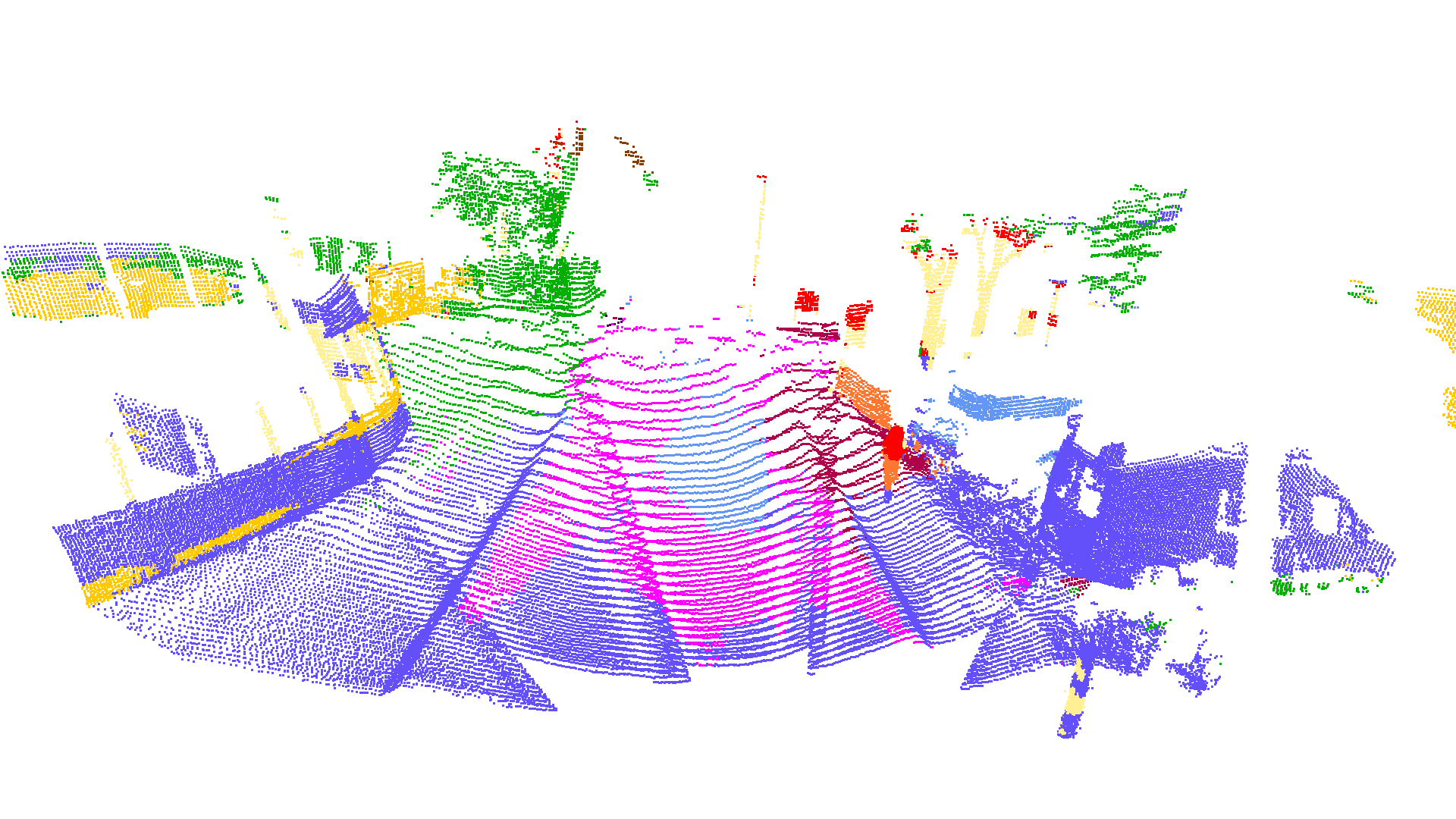}
        \caption*{Cfg. 1a: Pre-trained}
    \end{minipage}\hfill
    \begin{minipage}{0.325\textwidth}
        \centering
        \includegraphics[width=\linewidth]{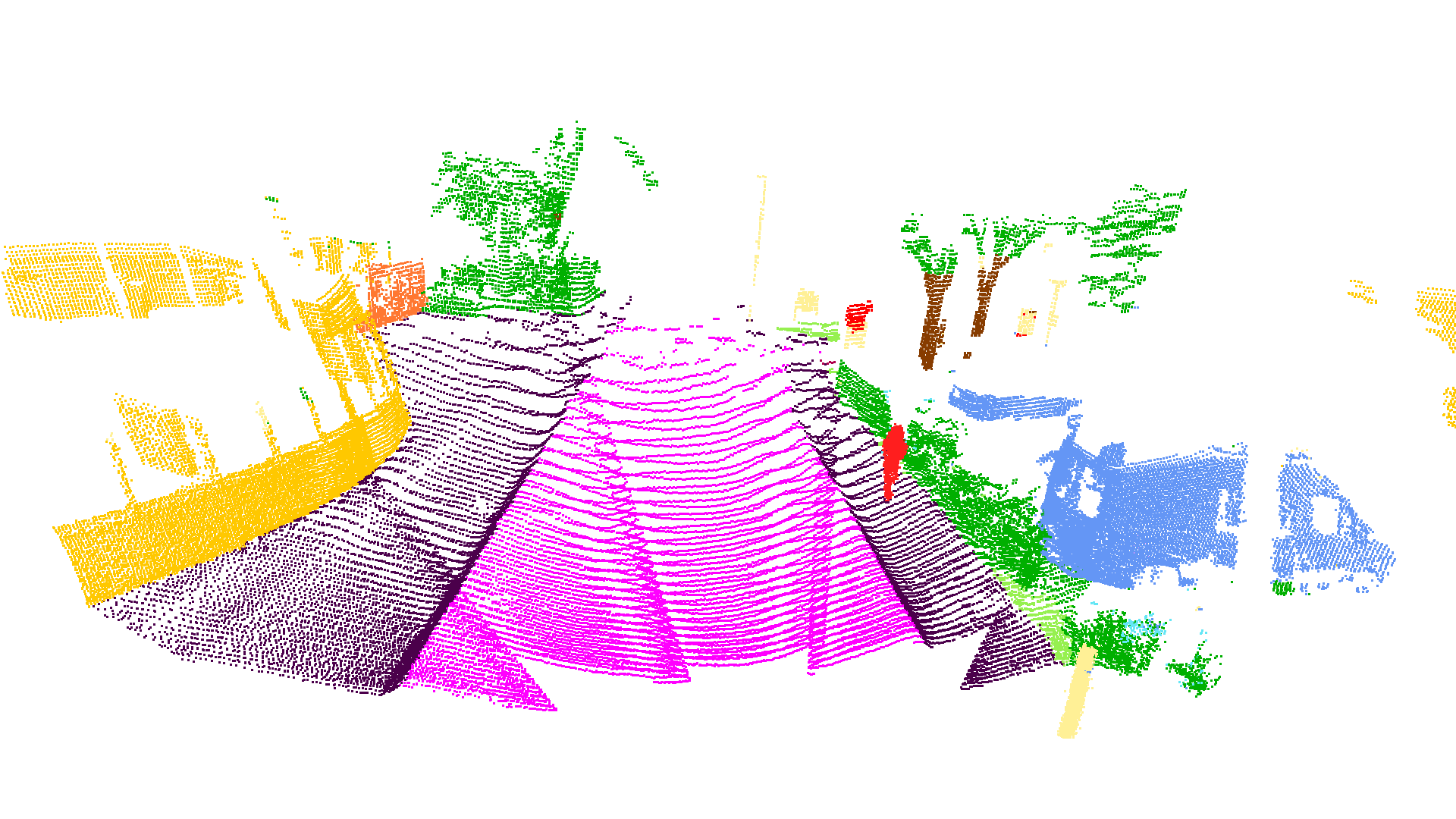}
        \caption*{Cfg. 2d: Fine-tuned}
    \end{minipage}\hfill
    \begin{minipage}{0.325\textwidth}
        \centering
        \includegraphics[width=\linewidth]{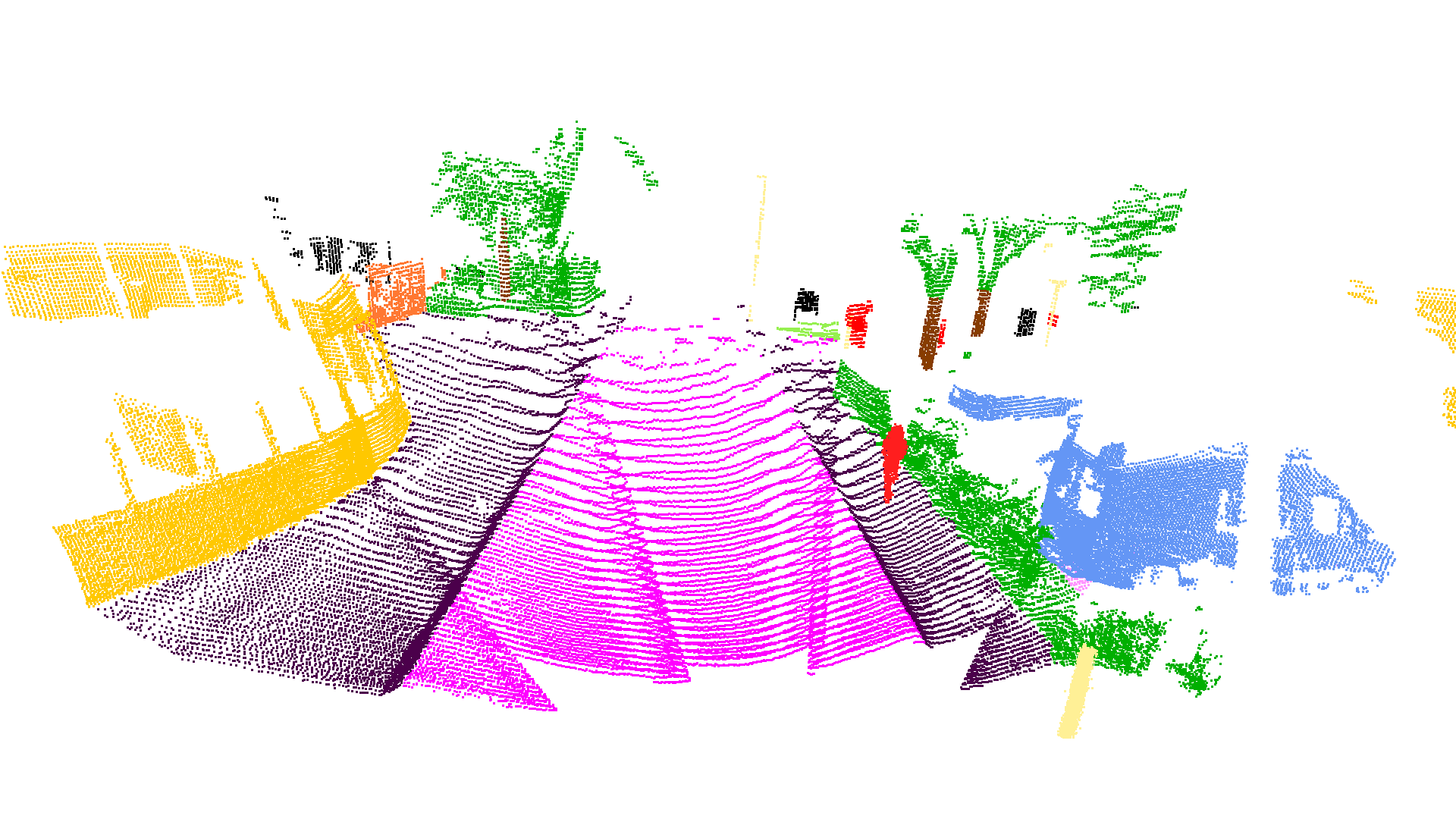} 
        \caption*{Ground Truth}
    \end{minipage}
    
    \begin{minipage}{0.32\textwidth}
        \centering
        \includegraphics[width=\linewidth]{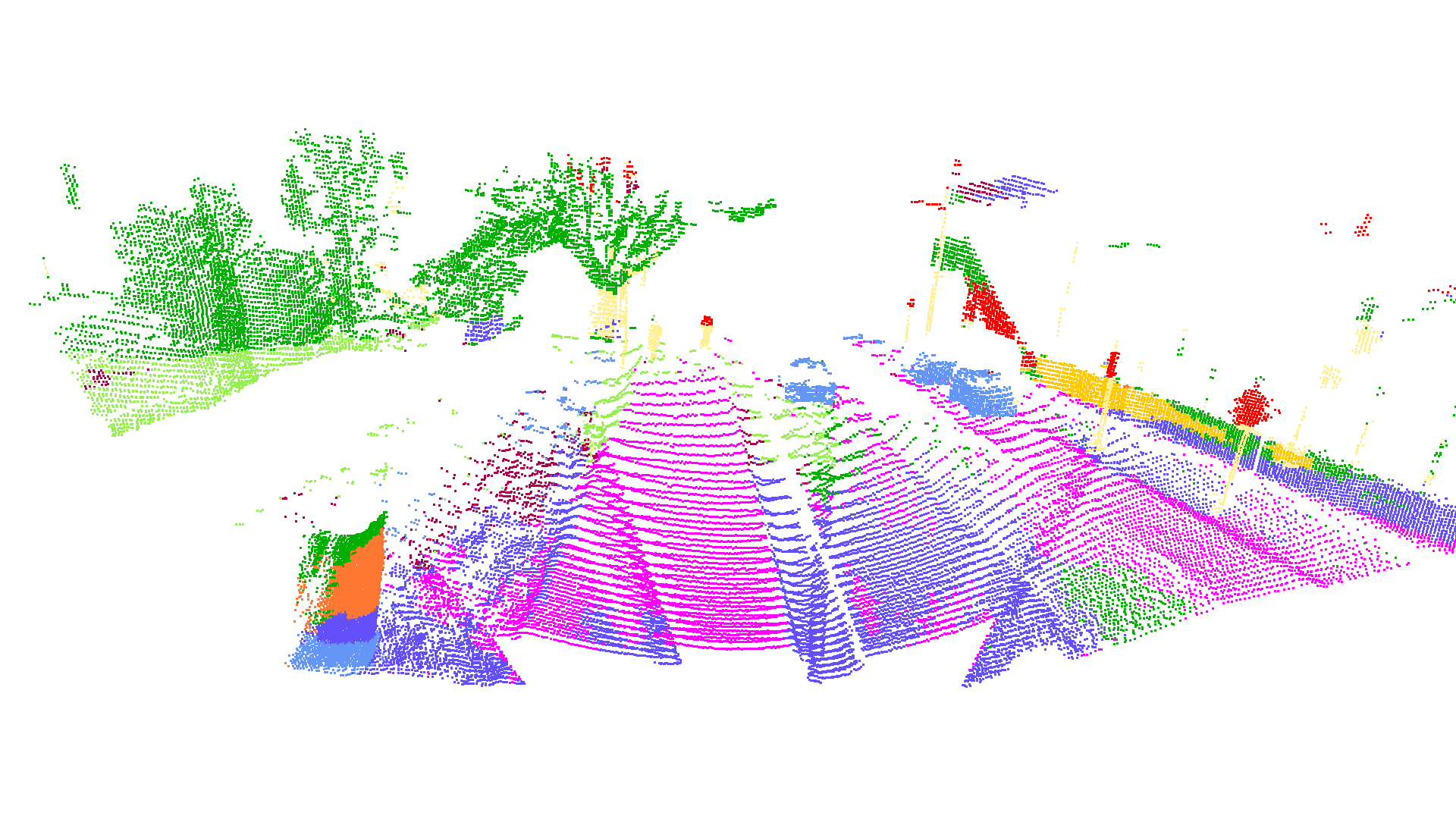}
        \caption*{Cfg. 1a: Pre-trained}
    \end{minipage}\hfill
    \begin{minipage}{0.32\textwidth}
        \centering
        \includegraphics[width=\linewidth]{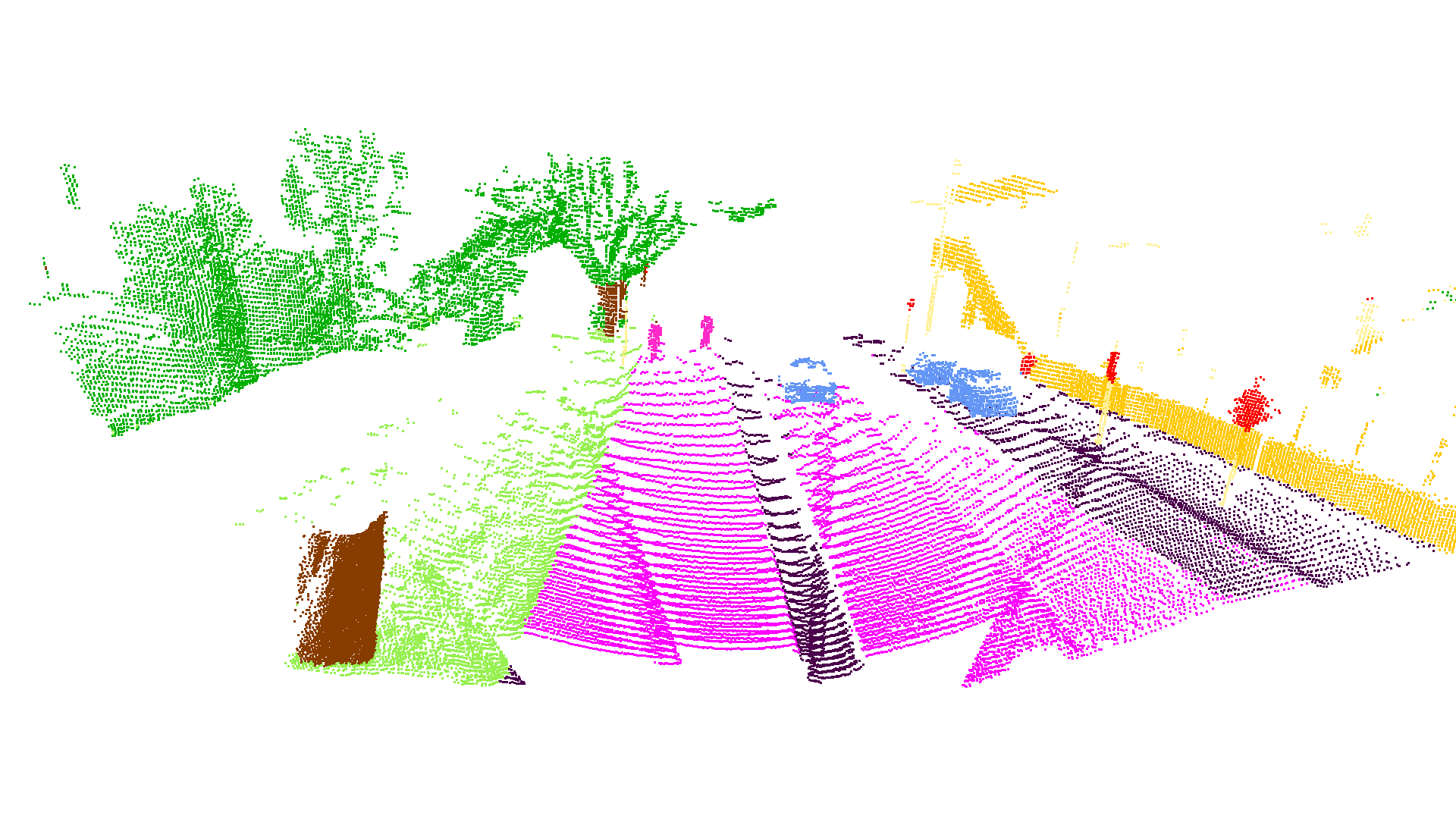}
        \caption*{Cfg. 2d: Fine-tuned}
    \end{minipage}\hfill
    \begin{minipage}{0.32\textwidth}
        \centering
        \includegraphics[width=\linewidth]{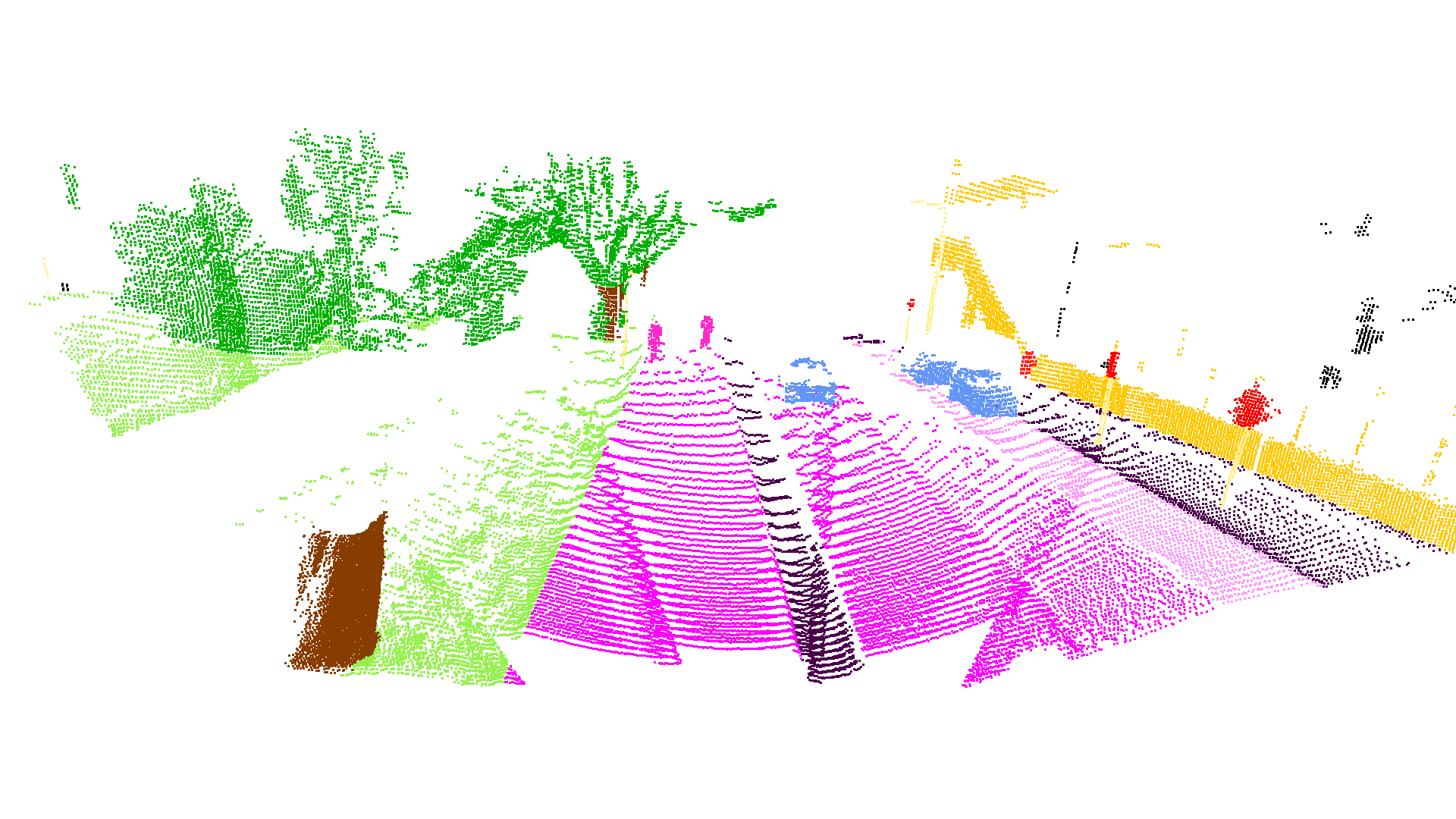} 
        \caption*{Ground Truth}
    \end{minipage}
    
    \caption{Labels predicted by the baseline Semantic\-KITTI pre-trained model (left) vs. the best performing fine-tuned model (middle) on two scenes of the test set of BikeScenes. The color scheme follows the Semantic\-KITTI palette. 
    \textcolor[RGB]{255,200,0}{\rule{0.8em}{0.8em}} building;  
    \textcolor[RGB]{255,0,255}{\rule{0.8em}{0.8em}} road;
    \textcolor[RGB]{75,0,75}{\rule{0.8em}{0.8em}} sidewalk;  
    \textcolor[RGB]{0,175,0}{\rule{0.8em}{0.8em}} vegetation;  
    \textcolor[RGB]{150,240,80}{\rule{0.8em}{0.8em}} terrain;
    \textcolor[RGB]{135,60,0}{\rule{0.8em}{0.8em}} trunk;
    \textcolor[RGB]{100,150,245}{\rule{0.8em}{0.8em}} car;
    \textcolor[RGB]{255,30,30}{\rule{0.8em}{0.8em}} person;
    \textcolor[RGB]{255,0,0}{\rule{0.8em}{0.8em}} traffic-sign;
    \textcolor[RGB]{255,40,200}{\rule{0.8em}{0.8em}} bicyclist;
    \textcolor[RGB]{0,0,0}{\rule{0.8em}{0.8em}} unlabeled/ignored.
    }
    \label{fig:model_predictions}
\end{figure}

Figure~\ref{fig:model_predictions} visualizes the predictions of the Semantic\-KITTI-only baseline (Cfg.~1a) and our selected model (Cfg.~2d) on two scenes of the BikeScenes test set. Cfg.~1a transfers poorly to BikeScenes: labels are largely wrong and inconsistent, which illustrates the shift between the pre-training data and the BikeScenes test data. By contrast, Cfg.~2d predicts coherent structures with sharp boundaries for common classes such as \emph{road}, \emph{building}, \emph{vegetation}, \emph{car}, and \emph{bicyclist}.

\FloatBarrier

%% file: Chapters/Conclusion.tex
\section{Conclusion}
\label{sec:conclusion}
Our work addresses the challenge of training 3D LiDAR semantic segmentation models for a real-world bicycle platform.

On the validation split, fine-tuning FRNet on BikeScenes outperforms training from scratch across matched schedules. On the test split, the validation-selected fine-tuned model achieves 63.6\% mIoU; the Semantic\-KITTI-only baseline achieves 13.8\%. The final model performs strongly on road, building, person, and bicyclist, while bicycle, motorcycle, and parking remain challenging.

BikeScenes currently covers one 1.3\,km closed-loop route, with train, validation, and test sets formed from distinct subsequences. Future work should add routes and regions and conduct matched cross-platform experiments to isolate the effects of sensor characteristics, mounting position, and scene context.

Our primary contributions are the introduction of BikeScenes-lidarseg, a bicycle-centric LiDAR semantic segmentation dataset, and an initial quantitative evaluation of transfer from automotive LiDAR data to a bicycle-mounted sensor setup. These results indicate that cyclist-centric perception benefits from dedicated data and benchmarks rather than direct transfer from automotive systems alone. We hope this work supports future research on bicycle safety systems and promotes bicycles as platforms for data collection and mapping purposes.